%% file: iclr2025_conference.tex
\documentclass{article} 
\usepackage{iclr2025_conference,times}

\input{math_commands.tex}

\usepackage{graphicx}
\usepackage{subcaption}
\usepackage{hyperref}
\usepackage{url}
\usepackage{tabularx}
\usepackage{booktabs}
\usepackage[T2A, T1]{fontenc}
\usepackage[utf8]{inputenc}
\usepackage{pdfpages}
\usepackage{capt-of}

\usepackage{wrapfig}
\title{BayesPrompt: human readable prompts that make sense}

\author{Franky Kevin Nando Tezoh, Ali Hussaini Umar, Alessandro Laio,  \& Guido Sanguinetti \thanks{Correspondence should be addressed to laio@sissa.it, gsangiun@sissa.it, fnandote@sissa.it.} \\
Department of Physics,\\
Scuola Internazionale Superiore di Studi
Avanzati (SISSA),\\
Trieste, Italy,\\
\texttt{\{fnandote,aumar,laio,gsanguin\}@sissa.it} \\
\And
Riccardo Rende \\
Center for Computational Quantum Physics, \\
Flatiron Institute, \\
162 5th Avenue, New York, NY 10010, \\
\texttt{\{rrende\}@flatironinstitute.org} \\
}

\iclrfinalcopy 
\begin{document}

\maketitle

\begin{abstract}

Reconstructing prompts that can elicit a desired answer or behaviour in an LLM is an open and important research topic.
Optimisation methods which aim at minimising the perplexity of a given answer, however, consistently yield  so-called \emph{pseudoprompts}, unintelligible strings of tokens which can lack  human interpretability. We argue that this is a consequence of the ill-posedness of the prompt optimisation task. By reframing the task as a Bayesian posterior inference over prompts, we propose an efficient algorithm to sample prompts which are both efficient (in terms of perplexity) and human readable. We compare our approach with state of the art alternatives showing on a real data set a marked improvement over a range of metrics.

\end{abstract}

\section{Introduction}

In-context learning (ICL)~\citep{dong2022survey} has recently become a major focus in machine learning. Researchers found that Large Language Models (LLMs) can solve diverse tasks like translating languages, analyzing sentiments, or solving math problems simply by conditioning  on a few exemplars  within a prompt. They do this without requiring any  update to the model's internal parameters. 
Much research has been dedicated to trying to explain what gives rise to the surprising in-context learning behaviour of LLMs \citep{rai2024practical,singh2024rethinking,vulic2020probing,jin2024exploring,ju2024large,elhage2021mathematical,olsson2022context}; regardless of its origin, the ICL phenomenon has also been immediately recognised for its practical importance. 

A key ingredient for ICL is designing the correct \emph{prompt} for a given task. Prompt optimisation approaches~\citep{sinha2021masked} aim to discover high confidence (low perplexity) prompts to leverage ICL for practical purposes, using both continuous soft prompts~\citep{lester2021power,liu2024gpt} and discrete hard prompts~\citep{wen2023hard} to construct more  effective inputs.

Many  studies have addressed the prompt-optimisation problem,  revealing a striking gap between human intuition and model behavior: optimal prompts are often short non-sensical sequences of tokens that nonetheless reliably steer a model towards a target output. Such {\it pseudoprompts} were first identified through universal adversarial triggers~\citep{wallace2019universal}; this phenomenon has since been shown to generalize, as models conditioned on unnatural, human-illegible sequences can match or exceed the performance of natural prompts~\citep{zou2023universal,kervadec2023unnatural,cherepanova2024talking}, consistent with evidence that models may rely more on distributional statistics than on human syntactic structure~\citep{sinha2021masked}. This suggests that the space of effective prompts is far larger, and far less human readable, than the space of prompts a person would ever write. Subsequent studies on Unnatural Language Processing  broadened this observation, challenging the assumption that LLMs rely on human-centric linguistic structures: ~\citep{zou2023universal,kervadec2023unnatural,cherepanova2024talking} demonstrated that models prompted with unnatural sequences which are semantically and syntactically incoherent to humans still produce predictable, meaningful outputs.

\paragraph{Contribution:} 
In this paper, we aim to illustrate the statistical origins of pseudoprompts and to propose an efficient algorithm that produces highly efficient prompts (in terms of resulting in the correct output with low perplexity) which are also human interpretable. Our motivation is both in terms of obtaining more readable, and hence presumably trustable, prompts, and to understand the operational modalities of LLMs when run to solve an inverse problem.

Our main contributions are:\begin{itemize}
    \item we reformulate the prompt optimisation problem in terms of Bayesian inference, showing that a neglect of the prior term gives rise to the pseudoprompt phenomenon;
    \item we propose an efficient algorithm, based on Markov-chain Monte Carlo, to obtain human readable  prompts;
    \item We illustrate how {\it reverse language modelling} can be used to provide an effective initialisation to both Bayesian and optimisation methods that greatly improves prompt reconstruction quality.
\end{itemize}

We validate both qualitatively and quantitatively the new framework on a range of question and answer tasks, showing that it obtains low perplexity prompts with high readability, significantly improving on existing techniques.

\section{RELATED WORK}

Learning a prompt from a given  answer can be framed as a reverse engineering problem: Large Language Models (LLMs) are trained to predict an answer given a question, and recovering the prompt requires inverting the direction. Prompt recovering can be formulated either in discrete token space or continuous embedding space. 
Among the earliest works on automatic prompt learning in  continuous space,~\citep{lester2021power,li2021prefix} showed that learned soft prompts yield higher confidence in the target answer than natural, human-interpretable prompts. However,~\citep{lester2021power} note that once decoded into discrete tokens, these learned prompts are gibberish, and incomprehensible to humans. To make such prompts interpretable upon decoding,~\citep{wen2023hard} alternate between projecting the current embedding sequence onto the vocabulary and computing the  gradient over the  projected embeddings to update the learned representations. Still, the resulting  prompts lack fluency.

In the discrete setting, gradient based discrete token search was first introduced by~\citep{ebrahimi2018hotflip}, who use  the gradient with respect to one-hot token encoding to identify the token replacement that most increases the classification loss. This  strategy was later adapted by~\citep{zou2023universal} to the setting of eliciting a specific target generation from an LLM. Building on this,~\citep{melamed2023prompts} showed that for a given ground-truth question, there exists a corresponding evil twin that elicits the same behavior as the ground-truth, and further showed that regularizing the objective function with the fluency term produces  prompts that are fluent but lack  similarity with the ground-truth question. Later,~\citep{rakotonirina2025evil} and~\citep{melamed2025demystifying} study the structure of these evil twins in more depth, finding that they contain a  mixed  language-like and non-linguistic tokens.



MCMC  sampling, a key ingredient in our approach, has been applied to text generation more broadly, though not to prompt learning itself.~\citep{goyal2021exposing} show that the masked conditionals of Masked Language Models (MLMs) do not correspond to a valid joint distribution, and instead use them as proposal distribution within a Metropolis-Hastings sampler to draw samples from the model's implicit energy. Reverse language models, another key ingredient, have been explored for purposes other than sampling:~\citep{yin2026ledom} train  a purely reverse language model from scratch and use it to construct a posterior based reward function, reranking forward generated outputs by how well the reconstruct the input. In our approach we use the reverse model as warm-start initialiser for the MCMC, and not for  post hoc reranking the prompts. To our knowledge, no prior work formulates prompt learning itself as a sampling problem.

In summary, existing literature presents a strict trade-off: optimised prompts are either highly effective but completely uninterpretable, or fluent but semantically disconnected from the target task. Critically, no prior method achieves both high task confidence and human fluency; the two properties are always in direct tension with one another. This is the gap our framework addresses.

\section{Bayesian Reformulation of the problem}\label{Meth}

Large language models (LLMs)  learn the conditional distribution of the subsequent tokens given a preceding context. Let $\left(\rva,\rvq\right) \sim  R$ be a pair of  answer and question (each defined by a sequence of tokens). 

%
%
%
%

Given a pretrained model one could learn the input/question tokens that minimize  the negative log-likelihood of the answer $\va$: 

\begin{equation} \label{opt}
    \min_{\vq \in \mathcal{V}^{n}} \mathcal{L}(\vq), \quad \text{where} \quad \mathcal{L}(\vq)= -\log P(\va|\vq),
\end{equation}

where  $\mathcal{V}$ denotes the model's vocabulary and $n$ is the sequence length of the question.  $P(\va|\vq)$ can be explicitly estimated by a LLM trained on next token prediction as $\prod_{i=1}^{m} P(\eva_i|\va_{<i},\vq) = \prod_{i=1}^{m}\mathrm{softmax}\big(f(\vq, \va_{<i})\big)_{\eva_i}$, where $\vz_i = f (\vq, \va_{<i})\in \mathbb{R}^{|\mathcal{V}|}$ is the logit vector the LLM outputs at position $i$, $m$ is the length of the answer $\va$. In this equation, the function $f$ is defined by the neural networks, and therefore depends on a (large) set of parameters, which in our derivation are considered as given and held fixed.  

However, this approach to learning the questions leads to prompts which are often ungrammatical sequences of seemingly random tokens. This stems from the ill-posed nature of \eqref{opt}. To ensure that \eqref{opt} is well defined, we must incorporate a prior distribution over the questions $P(\vq) = \prod_{i=1}^{n} P\left(q_i|\vq_{<i}\right)$.  Consequently, our objective becomes:

\begin{equation} \label{opt1}
    \min_{\vq \in \mathcal{V}^{n}} \mathcal{L}(\vq), \quad \text{where} \quad \mathcal{L}(\vq)= -\log P(\va|\vq)-\log P(\vq).
\end{equation}


By incorporating the prior distribution over the inputs $P(\vq)$, we encourage  the optimisation framework to prioritize  the generation of fluent  prompts.  To solve the optimisation problem in \eqref{opt1}, we use two algorithms described in the literature, greedy coordinate gradient (GCG) and gradient descend (GD), and the approach which we propose, Markov Chain Monte Carlo (MCMC).

\paragraph{Greedy coordinate Gradient (GCG)}~\citep{zou2023universal}: Gradient information is used to get potential token replacements for a given position by evaluating $\displaystyle -\nabla_{\ve_{q_i}}\mathcal{L}(\vq)$ with respect to one-hot vector $\ve_{q_i}$ of the token at the position $i$; then we select the top-$k$ tokens which maximize the gradient of $-\mathcal{L}(\vq)$, $\mathcal{X}_i := \text{Top-}k\left( -\nabla_{\ve_{q_i}}\mathcal{L}(\vq)\right)$. We construct  $B$ proposals sequence tokens $\{\tilde {\vq}^{(b)}\}_{b=1}^{B}$, by  selecting a position $i$ uniformly at random and replacing the original token with a candidate $\tilde{\evq}_i^{(b)} \sim \text{Uniform}(\mathcal{X}_i)$. Finally, the best proposal is selected greedily based on the actual loss value.  This process is iterated for a fixed number of steps or until convergence.

\paragraph{Gradient Descent (GD):} It involves  applying gradient descent on the sequence embedding $\vq_{emb}\in \mathbb{R}^{n\times d}$ to minimize the loss $\mathcal{L}(\vq_{emb})$, where $d$ is the embedding dimension. After optimising the continuous embedding space, the final sequence is  obtained by mapping the learned embedding back to the nearest discrete tokens in the vocabulary tokens. A natural extension of this approach, known as Hard prompts made EaZy, in abbreviation PEZ~\citep{wen2023hard} interleaves this nearest-neighbor embedding assignment with the gradient updates: at each iteration, the current embeddings are first replaced by their closest vocabulary embeddings via cosine similarity, and the gradient is then computed at those vocabulary-aligned embeddings before the optimiser update is applied. This ensures that the gradient signal is always evaluated at an embedding that belongs to the vocabulary embedding matrix, yielding a more principled optimisation than standard GD.

\subsection{Reverse Language Model}\label{rlm1}

Standard autoregressive language models are trained to sequentially predict each token in a sequence given all the previous observed tokens. Formally, given a sequence of $n$ tokens $\vq = \left(\evq_1,\evq_2,\dots,\evq_n\right)$,  the joint probability of the sequence admits the following factorization:

\begin{equation}\label{fwd}
    P(\vq) = \prod_{i=1}^{n}P(\evq_i|\vq_{<i}),
\end{equation}

where $\vq_{<i} = \left(\evq_1,\evq_2,\dots, \evq_{i-1}\right)$ denotes the subsequence of all tokens preceding position $i$. Under this formulation, the model processes tokens strictly from left to right, and each conditional distribution $P(\evq_i|\vq_{<i})$ is parameterized by a neural network, typically a Transformer architecture~\citep{vaswani2017attention}. We refer to this conventional formulation as the \emph{forward model}.

A natural question arises: what if the sequence dependency is reversed? Rather than conditioning each token to its left context, one can construct a model that  processes the sequence from right to left. Concretely, given the original token sequence $\vq$, we define its reversal as:

\begin{equation} \label{reversal_token}
    \vq_{\text{rev}} = \left(\evq_n, \evq_{n-1},\evq_{n-2},\dots, \evq_1\right),
\end{equation}

obtaining by permuting the tokens indices in reverse order. A  \emph{reversed} or \emph{backward}~\citep{peters2018deep,yin2026ledom} model is then trained autoregressively on this reversed sequence, learning the following factorization:

\begin{equation}\label{bwd}
    P(\vq_{\text{rev}}) = \prod_{i=1}^{n}P\left(\evq_{n+1-i}|\vq_{\text{rev},<i}\right).
\end{equation}
where $\vq_{\text{rev},<i} = \left(\evq_n,\evq_{n-1},\dots, \evq_{n+2-i}\right)$ denotes the suffix context acting as the casual history for the target token $\evq_{n+1-i}$.

Importantly, the reversed model is not a post-hoc transformation of the forward model; it is an independently trained model whose training data consists of token sequences presented in reversed order. Consequently, the backward model learns a distinct set of attention patterns and internal representations shaped by right-to-left causal dynamics. 

\subsection{Metropolis-Hastings Algorithm.}

Our approach aims at  obtaining samples from the posterior distribution $P(\rvq\vert\rva)$. We do so using the classical Metropolis-Hastings (MH) algorithm, treating $-\log P(\rva,\rvq)$ term from \eqref{opt1} as target distribution. MH allows exploring the prompt space $\mathcal{V}^n$ in a way that respects the prior $\mathbb{P}(\rvq)$, ensuring natural language structure while still gravitating toward regions that maximize the likelihood of the answer $\va$ conditioned on the question $\vq$. 
 At each time step, we draw a new sample $\vq'$ from a proposal distribution $Q\left(\vq'| \vq\right)$, whose form we specify below, and we either accept or reject it with an acceptance probability equal to:
\begin{equation}
    \alpha\left(\vq',\vq\right) = \min \left(1, \dfrac{\pi(\vq') Q(\vq| \vq')}{\pi(\vq) Q(\vq'| \vq)}\right),
\end{equation}
where $\pi \left(\rvq\right)= P(\rva,\rvq)$.

The distribution of the sequence of  sample generated $\vq_{n}$ converges to the target distribution $\pi$ for irreducible and aperiodic Markov chains as $n$ goes to infinity. While the convergence to the stationary distribution of the algorithm is satisfied for any valid proposal and initial state $\vq_0$, practical deployment of MH requires an effective initialisation and a proposal distribution that promotes good mixing. 
\paragraph{Initialisation:} We initialise the process by sampling a question from a simple reverse model, obtained by finetunining the forward model on the reverse task. This avoids the high computational costs of evolving the MCMC from an initial state which is atypical for the distribution that we want to sample. This warm-start initialisation ensures the chain begins  with a linguistically fluent question, even if the initial confidence on the answer is relatively low.


\paragraph{Proposal Distribution for the MH sampler:}
Transitioning from state $\vq$ to state $\vq'$ involves a set of discrete edit operations, specifically replacement, insertion, and deletion. While replacement modifies one or more tokens within the existing sequence, insertion and deletion operations dynamically adjust the  sequence length.

\paragraph{Actions:} Let $\vq = \left(\evq_1,\evq_2,\dots,\evq_{i-1},\evq_i,\evq_{i+1},\dots,\evq_n\right)$ be the current sequence of question tokens. To generate a proposal state $\vq'$, we select a token index $i\in \{1,\dots,n\}$ uniformly at random and replace $\evq_i$ with a new token $\evq'_i$, sampling from the probability distribution generated by either the left ($\vq_{<i}$) or the right context ($\vq_{>i}$). The proposal probability $Q(\vq'| \vq)$ is then defined by the selection of either the forward or the reverse model:

\begin{equation}
    Q(\vq'| \vq)= \frac{1}{n}\times
    \begin{cases}
    P(\evq'_i| \vq_{<i}) & \text{ if using the forward model}\\
    P_{\text{rev}}(\evq'_i|\vq_{>i},\va) & \text{if using the reverse model}
    \end{cases}
\end{equation}

This formulation ensures that the local edit is informed by either the preceding or succeeding  lexical context.

To perform an insertion~\citep{miao2019cgmh}, we select an index $i\in \{1,\dots,n+1\}$ uniformly at random to determine the position of  a new token $\evq^*$. This results in an expanded sequence of length $n+1$, $\vq=\left(\evq_1,\evq_2,\dots,\evq^*,\evq_i,\evq_{i+1},\dots,\evq_n\right)$. The proposal probability is defined by the selection of the position and the sampling of the new token:

\begin{equation}
    Q(\vq'| \vq)= \frac{1}{n+1}\times
    \begin{cases}
    P(\evq^*|\vq_{1:i}) & \text{ if using the forward model}\\
    P_{\text{rev}}(\evq^* | \vq_{i+1:n},\va) & \text{if using the reverse model}
    \end{cases}
\end{equation}

Lastly, to perform deletion, we select an index $i \in \{1,\dots,n\}$ uniformly at random and remove the corresponding tokens $\evq_i$. This operation reduces the sequence length from $n$ to $n-1$, resulting in the state $\vq'=(\evq_1,\evq_2,\dots,\evq_{i-1},\evq_{i+1},\dots,\evq_n)$. Since this operation involves only the selection of the position and no token sampling, 

The proposal probability is defined simply as:

\begin{equation}
    Q(\vq'|\vq)= \frac{1}{n}.
\end{equation}

This  ensures that any token within the sequence has an equal probability of being targeted for deletion during the Markov chain transition.

\section{Implementation details}

\paragraph{Dataset and Model:} We evaluate our methodology using the NQ-OPEN dataset, an open domain benchmark structured into natural language question-answer pairs, using the standard train/test split. This configuration enables the model to establish direct semantic mappings between queries and their respective targets. We fine-tune  Llama-3.2-1B-Instruct model  on the NQ-OPEN \textsc{train} split in a supervise manner to optimize its generative performance~\citep{shani2025tokens}.

\paragraph{GCG and GD for NQ-OPEN \textsc{test}:} Given the forward model Llama-3.2-1B-Instruct fine-tuned on NQ-OPEN dataset~\citep{huang2025llm}, we perform prompt learning on \eqref{opt1} for each answer in NQ-OPEN \textsc{test}, using  optimisation algorithms such as greedy coordinate gradient for discrete token selection and gradient descent for learning in continuous space described.

\paragraph{Reverse-model:} We finetune Llama-3.2-1B-Instruct model with Low-Rank Adaptation (LoRA)~\citep{hu2022lora} on NQ-OPEN \textsc{train} split sequences whose token order is fully reversed, so the model learns to generate a question, in reverse, when prompted with a reversed answer (see section~\ref{rlm1}).


\paragraph{MCMC for NQ-OPEN \textsc{test}:} Following the fine-tuning of both the forward and reverse Llama-3.2-1B-Instruct models on NQ-OPEN dataset, we employed the Metropolis Hastings algorithm to explore the posterior distribution of the input given the answer $P\left(\vq |\va\right)\propto P\left(\va |\vq\right) P\left(\vq\right) $. This MCMC approach allows us to generate a diverse set of question samples conditioned on a given answer. For each MCMC trajectory under a fixed answer, we retain exclusively the final state of the Markov chain as the optimised question. The transition proposal distribution relies on token level mutations with action probabilities set to $p_{\text{replace}} = 0.6$, $p_{\text{delete}} =0.2 $, and $p_{\text{insert}}= 0.2$ (ensuring $\sum p = 1$). Specially, a replacement move modifies at most two tokens simultaneously, whereas insertion and deletion moves operate on a single token per step. We provided a detailed statistical analysis of the  MCMC sampling process to illustrate its convergence and efficiency in the prompt learning task in Appendix~\ref{conver}.

\section{Results and Discussion}

This section reports the primary evaluation metrics for the learned prompts, comparing the optimisation-based baselines (GCG, GCG-Reg and GD-PEZ) against our proposed MCMC approach, which uses the forward model to generate proposals. We analyze two quantities: the fluency $P(\vq)$ and the confidence, defined as the conditional probability of the answer given the question $P(\va|\vq)$. Beyond comparing MCMC to the baselines directly, we also examine how much each baseline benefits from reverse-model warm-start initialisation relative to random initialisation. The resulting performance gap between the two schemes directly quantifies the importance of a good initialisation for the prompt inversion problem, a non trivial consideration given the vast, discrete, and highly non-convex nature of the token search space.

\subsection{Distributional-Level Analysis of the Methods}\label{MH_test}

\begin{figure}[!h]
\centering
\begin{subfigure}[b]{0.48\textwidth}
\centering
\includegraphics[width=\textwidth]{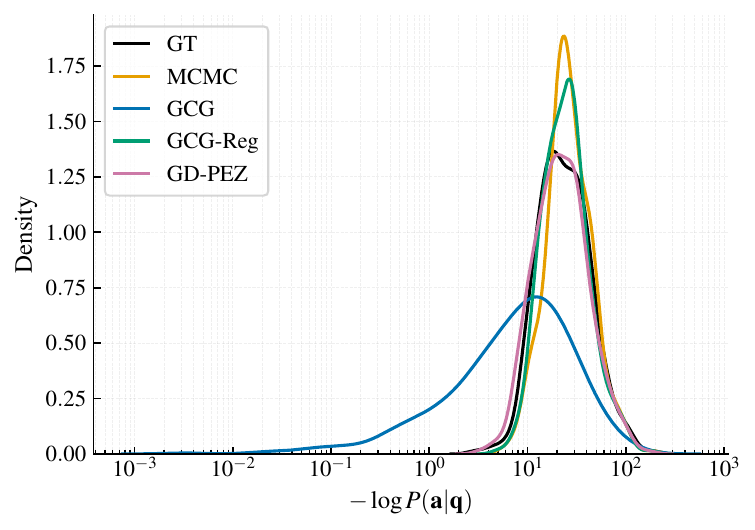} 
\caption{}
\label{gd_prob_grid}
\end{subfigure}
\hfill
\begin{subfigure}[b]{0.48\textwidth}
\centering
\includegraphics[width=\textwidth]{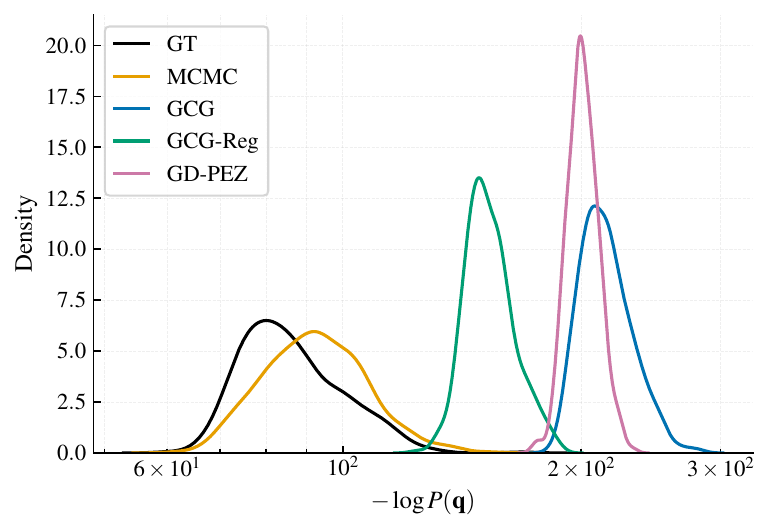} 
\caption{}
\label{gd_lik_grid}
\end{subfigure}
\caption{Comparison of the distribution of probability between the ground truth (black), the optimisation-based methods (GCG, GCG-Reg, GD-PEZ) under random initialisation and the MCMC-sampled approach. Random initialisation results are computed over $371$ question-answer pairs from the NQ-OPEN dataset, using a prompt length of $10$ tokens. \textbf{Left:} Distribution of the negative log-conditional probability $-\log P(\va| \vq)$, measuring how well the generated question recovers the target answer. \textbf{Right:} Distribution of the negative log-fluency, $-\log P(\vq)$, measuring the linguistic naturalness of the generated prompt.}
\label{gcg_grid}
\end{figure}


 As shown in Figure~\ref{gd_prob_grid}, GCG-Reg's and GD-PEZ's confidence distribution under random initialisation both closely tracks the ground-truth (GT) distribution, while MCMC's confidence is modestly lower than GT, with a distribution notably more concentrated (peaked) around its mode than the others. In contrast, GCG under random initialisation exhibits markedly higher confidence than GT, with a pronounced  heavy tail toward much higher confidence values, indicating that a substantial fraction of its recovered prompts are far more confident in the answer than GT or any other method, consistent with unregularized GCG over-optimizing for confidence in the answer. Figure~\ref{gd_lik_grid} shows that MCMC achieves the highest fluency among all methods, with a distribution closest to the ground truth, while GCG, GCG-Reg, and GD-PEZ under random initialisation all remain noticeably less fluent.  Taken together, these results confirm that MCMC is the only method to jointly achieve near GT confidence and the best fluency among the methods compared. GCG's excess confidence comes at the cost of the marked fluency gap seen in Figure~\ref{gd_lik_grid}. We refer to prompts that achieve confidence in the answer at least as high as GT  at the cost of near-zero fluency as pseudoprompts, a designation that captures this pathological trade-off between confidence and fluency more precisely and quantitatively than a purely qualitative description will allow. Hence, GCG, GCG-Reg, and GD-PEZ under random initialisation all produce pseudoprompts, whereas MCMC's confidence remains slightly below GT while achieving the best fluency among all methods compared, placing it outside this pathological trade-off.

\subsection{The Effect of Initialisation on Optimisation-Based Methods}

\begin{figure} [t ]
    \centering
    \includegraphics[width=1.0\linewidth]{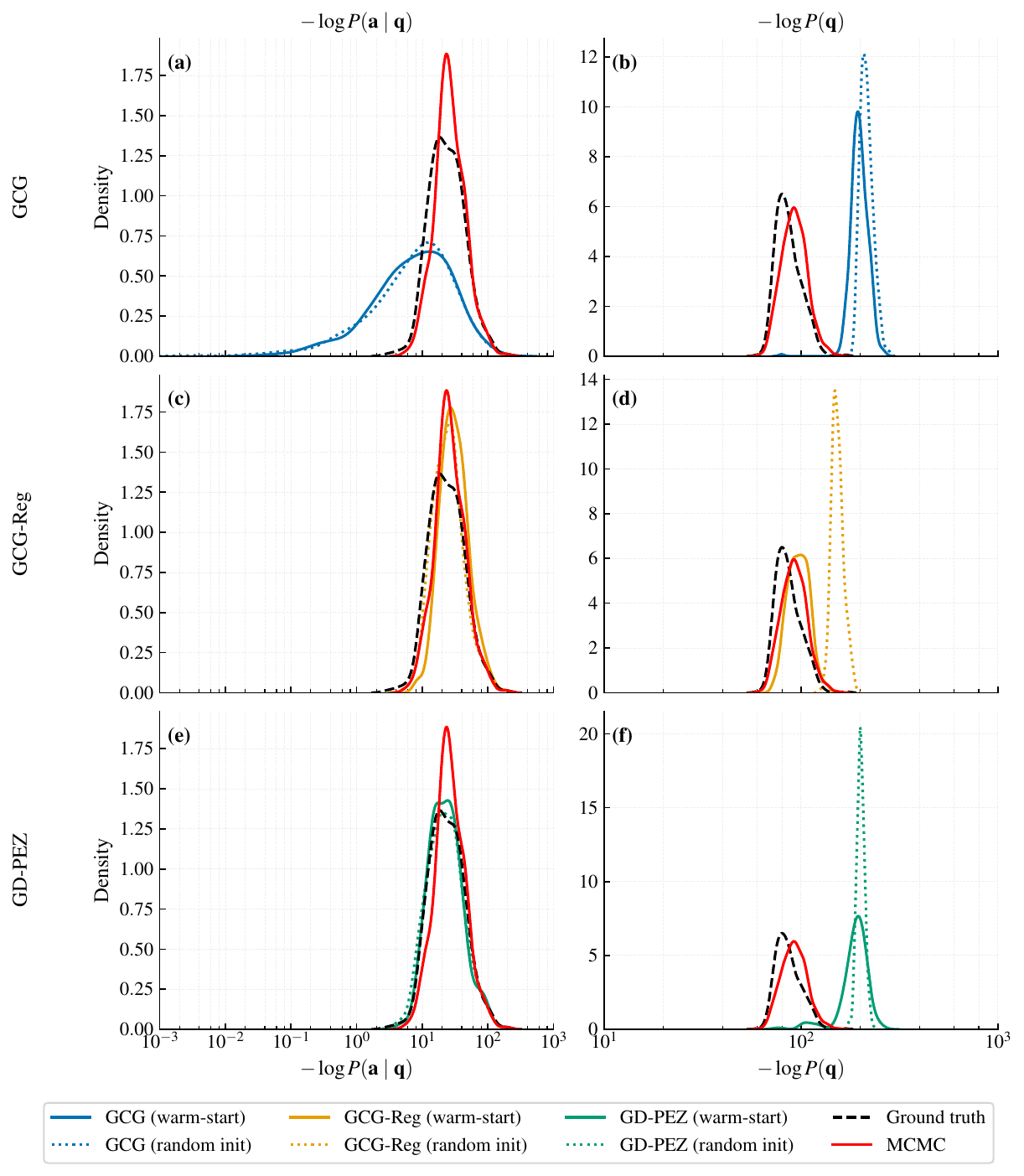}
    \caption{Effect of initialisation on optimisation-based methods (GCG, GCG-Reg, GD-PEZ), compared against ground truth (black dashed) and MCMC (red), evaluated over $371$ question-answer pairs fron NQ-OPEN dataset. Each row shows one method under warm-start (solid) and random (dotted) initialisation, with the same ground truth and the MCMC distributions repeated across rows for reference. \textbf{Left column (a,c,e):} Distributionod the negative log-conditional probability $-\log P(\va|\vq)$, measuring confidence in the recovered answer. \textbf{Right column (a,c,e):} Distribution of the negative log-fluency $-\log P(\vq)$, measuring the linguistic naturalness of the recovered prompt.}
    \label{fig:gcg_vs_}
\end{figure}

The effect of the initialisation on optimisation-based methods (GCG, GCG-Reg and GD-PEZ) is depicted by Figure~\ref{fig:gcg_vs_}. Warm-start initialisation does not meaningfully change GCG's confidence pattern: it remains substantially more confident than the GT under either initialisation, consistent with the heavy-tail behavior noted earlier. GD-PEZ, by contrast, shows confidence close to GT under both initialisation schemes, with warm-start and random-init curves nearly overlapping. GCG-Reg shows the opposite shift: while its confidence closely tracked GT under random initialisation, warm-start pushes it to become noticeably less confident than GT. Turning to fluency, warm-start initialisation yields a substantial improvement for all three methods relative to random initialisation, most dramatically for GCG-Reg, whose fluency distribution moves from being almost entirely separated from GT under random initialisation to a substantially overlapping with the GT under warm-start, though a distinguishable gap remains. GD-PEZ and GCG also improve under warm-start but remain far less fluent than GT. These results show the effect of initialisation is asymmetric across confidence and fluency: warm-start consistently and substantially improves fluency across all three methods, whereas its effect on confidence is method-specific, leaving GCG and GD-PEZ unchanged, and pushing GCG-Reg past GT in the opposite direction. Comparing all three optimisation-based baselines directly to MCMC, only GCG-Reg under warm-start approaches MCMC's performance in both confidence and fluency: its gap from the GT on both metrics is comparable in size to MCMC's, whereas GCG and GD-PEZ remain far more separated from GT on at least one of the two metrics even under warm-start. MCMC nonetheless remains closer to GT than GCG-Reg on both confidence and fluency, but GCG-Reg is the only baseline for which warm-start narrows the gap enough to be broadly comparable to MCMC, rather than dominated by it.

\subsection{Qualitative Analysis}


\begin{figure} [t ]
    \centering
    \includegraphics[width=1.0\linewidth]{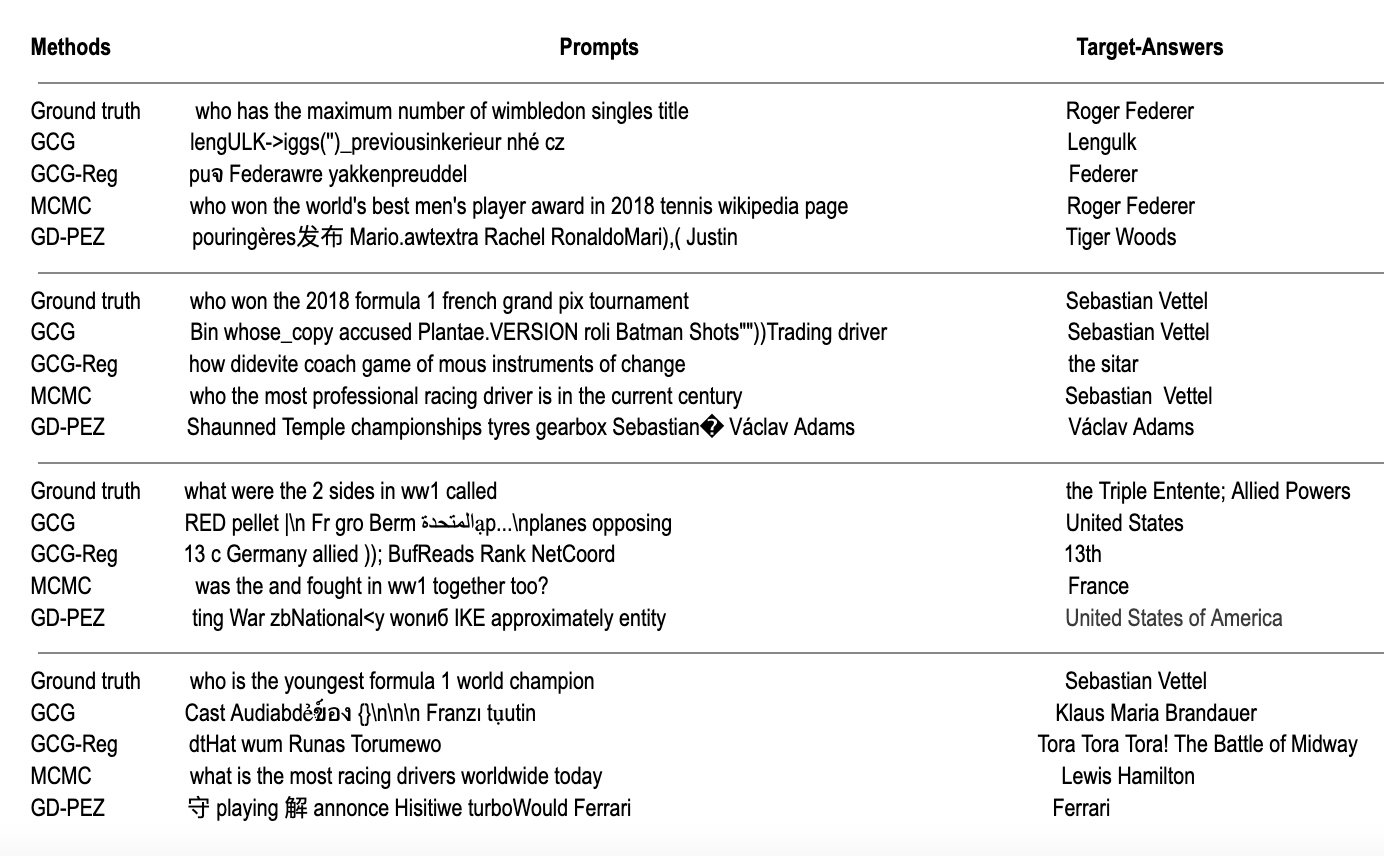}
    \caption{\textbf{Qualitative analysis of prompts learned  by the  optimisation-based methods and the sampling-based MCMC framework}. For each target answer, the ground truth question is shown alongside the corresponding  question recovered by GCG, GCG-Reg, and GD-PEZ  under random initialisation, and by our MCMC based sampler (final sample and forward proposal).}
    \label{fig:seman_res}
\end{figure}

\begin{figure} [!t]
    \centering
    \includegraphics[width=1.0\linewidth]{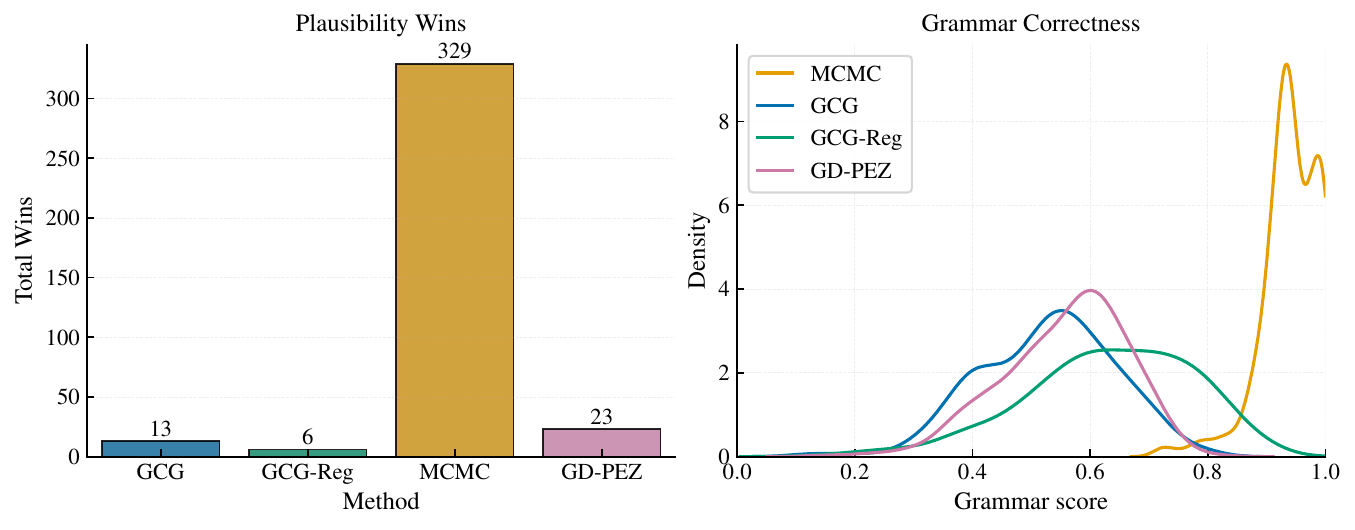}
    \caption{
 \textbf{Left:} Histogram showing the total distribution of winning instances for question-answer plausibility across  optimisation-based methods under random initialisation (GCG, GCG-Reg, GD-PEZ) and sampling framework (reporting the forward variant proposal for our MCMC approach ).  \textbf{Right:} Grammar correctness score distribution for the same evaluated methods.}
    \label{chat}
\end{figure}


Figure~\ref{fig:seman_res} provides qualitative examples of the prompts learned by optimisation-based methods  (under random initialisation) and the sampling-based MCMC framework. As observed, the candidate prompts optimised via unregularized GCG, GCG-Reg and GD-PEZ consist primarily of gibberish,  incorporating unnecessary formatting symbols alongside ungrammatical, mixed-language text~\citep{rakotonirina2025evil,melamed2025demystifying} that lacks grammatical, syntactic and semantic coherence. In contrast, the candidate prompts sampled via MCMC demonstrate clear human  readability and semantic interpretability, preserving structured linguistic features from the original ground-truth reference questions. Extensive examples of the learned prompts under random initialisation and the sampled MCMC prompts are provided in Appendix~\ref{qa_}, Figure~\ref{fig:seman_res_general}, and qualitative results on the optimisation-based methods under warm-start initialisation, alongside our MCMC approach are also depicted in Appendix~\ref{qa_}, Figure~\ref{fig:seman_res_}

To automatically assess the optimised/sampled question-answer pairs, Figure~\ref{chat} presents two metrics scored using an LLM-judge evaluation workflow modeled on LMSYS's (Large Model Systems Organization) Chatbot Arena (prompts detailed in the  appendix~\ref{plausibility_score}): the plausibility of the question-answer pair, and the grammar correctness of the question alone. Across both metrics, our MCMC approach for prompt learning comes out ahead of the baseline methods (GCG, GCG-REG and GD-PEZ under random initialisation), with MCMC showing the largest margin on plausibility ($329$ wins vs. $6$ for GCG-Reg, $13$ for GCG and $23$ for GD-PEZ), and a similarly clear separation on grammar correctness, where MCMC's scores concentrate sharply near $0.9$ -$1.0$, while GCG, GCG-Reg, and GD-PEZ all show broad distributions peaking in the $0.4$-$0.5$ range, with GCG-Reg shifted slightly further right than GCG and GD-PEZ. 



\section{Conclusion}


In this work, we recast prompt optimisation as a problem of Bayesian inference. We show that incorporating the prior over the question is essential to learn prompts that remain fluent rather than degenerating into uninterpretable token sequences. In contrast to prompt optimisation techniques such as  unregularized GCG and GD-PEZ, our approach, combined with the warm-start initialisation from a reverse model, learns prompts that achieve a genuine balance between answer confidence and fluency. While regularized GCG (GCG-Reg) with the prior distribution over the question and warm-start initialisation also improves this balance, our MCMC-based approach achieves slightly better confidence and fluency jointly. Moreover, and very importantly, our approach is the only one which, for a given target answer, allows sampling a \emph{probability distribution} of the questions, consistently with probabilistic formulation of language models (see eq. \ref{fwd}). 

Several directions are worth exploring. For the sake of computational efficiency all the analysis presented in this work has been performed on Llama-3.2-1B, a relatively small model. All the scheme can be ported with no modification to much larger models.  A second direction  is replacing Metropolis-Hastings with alternative sampling schemes, such as Rosenbluth Sequential Monte Carlo. A final intriguing direction is to extend prompt inversion to world models such as joint embedding predictive architecture (JEPA)~\citep{assran2023self}. Unlike autoregressive models,  JEPA learns by predicting abstract latent representations of its target input. It would be interesting to define a  notion of prompt inversion in a purely representational space, where the target is no longer a token sequence but a latent state. 

\section{Acknowledgements}

The authors would like to thank  Arsene Gibbs Nwemadji Tiako and Nafack Baurice for insightful discussions and suggestions.

\bibliography{iclr2025_conference}
\bibliographystyle{iclr2025_conference}

\newpage
\appendix

\section{Comparative graphs} \label{comparative_fig}

\begin{figure}[t]
\centering
\begin{subfigure}[b]{0.48\textwidth}
\centering
\includegraphics[width=\textwidth]{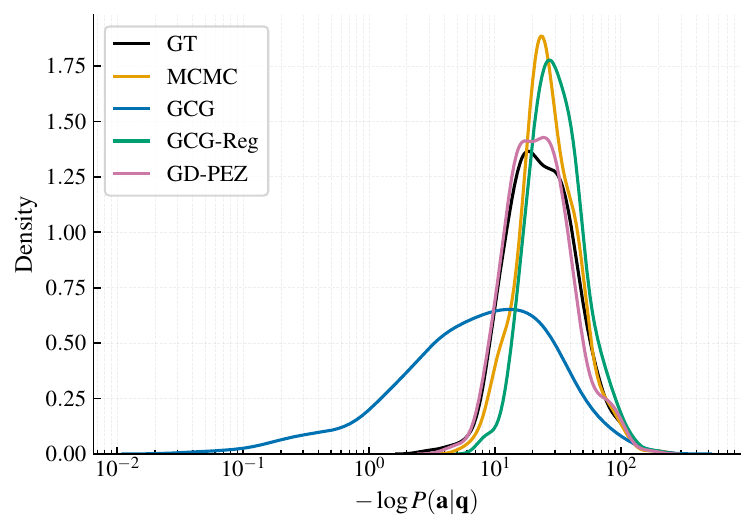} 
\label{compare_prob}
\end{subfigure}
\hfill
\begin{subfigure}[b]{0.48\textwidth}
\centering
\includegraphics[width=\textwidth]{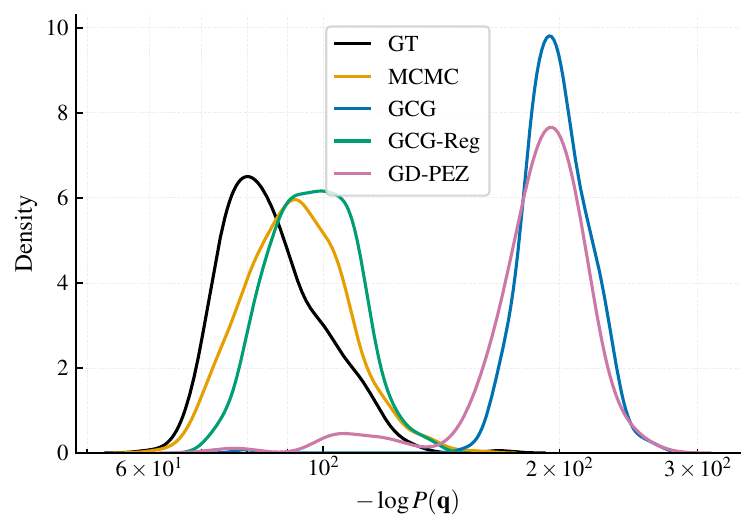} 
\label{comapre_lik}
\end{subfigure}
\caption{Comparison of the distribution of probability between the ground truth (blue), the sampling and  optimisation frameworks evaluated over  $371$ question-answer pairs under warm initialisation. \textbf{Left:} Distribution of the negative log-conditional probability $-\log P(\rva| \rvq)$. \textbf{Right:} Distribution of the negative log-fluency, $-\log P(\rvq)$. }
\label{compare_fram}
\end{figure}

Figures~\ref{compare_fram} present a comparative graph between the ground-truth, the sampling and the optimisation frameworks across all methods, under warm-start initialisation. These results show that MCMC is the only method that consistently yields prompts combining strong  fluency with high confidence in the target answer. Meanwhile, if the priority is  high confidence in the answer relative to the ground truth, GCG is the preferable choice.

\section{Convergence of the Metropolis-Hastings} \label{conver}

\begin{figure}[t]
\centering
\begin{subfigure}[b]{0.48\textwidth}
\centering
\includegraphics[width=\textwidth]{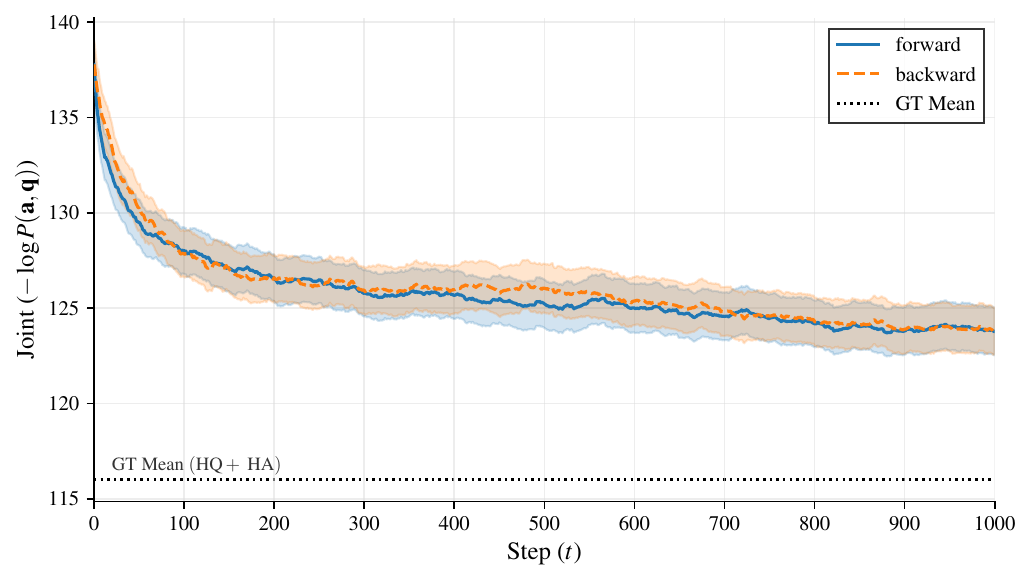} 
\end{subfigure}
\hfill
\begin{subfigure}[b]{0.38\textwidth}
\centering
\includegraphics[width=\textwidth]{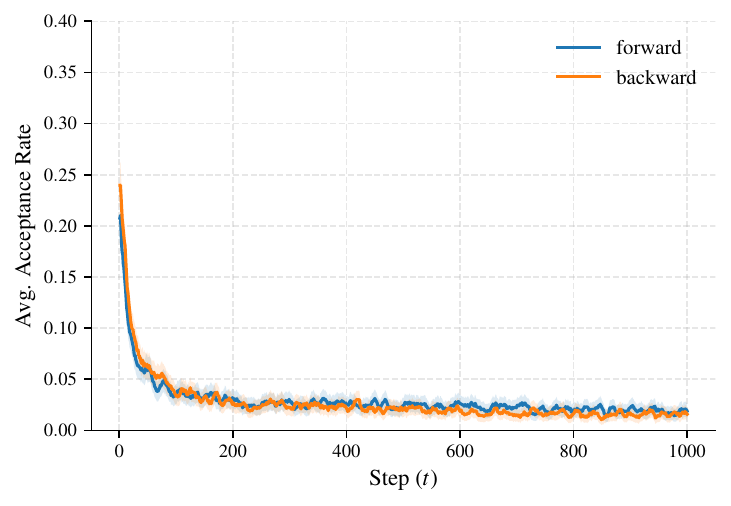} 
\end{subfigure}
\caption{\textbf{Left:} Mean over $371$ questions of the $-\log P(\va,\vq)$ as function of the number of steps for the forward and backward/reverse proposals under warm-start initialisation. \textbf{Right:} Mean acceptance rate as function of the number of steps for the same configurations.}
\label{mcmc_tr}
\end{figure}

Figure~\ref{mcmc_tr} (Left) shows the empirical mean of the joint negative log-probabilities $-\log P(\va,\vq)$ as function of the sampling step ($t$), computed as:

\begin{equation*}
    \E_{(\rva,\rvq)\sim R}\left[-\log P(\rva,\rvq)\right] \approx-\frac{1}{M}\sum_{i=1}^{M}\log P(\va_i,\vq_i).
\end{equation*}
We compare the forward and backward proposal strategies, both under warm-start initialisation. The warm started forward and backward chains converge to the same region of the target distribution, demonstrating that the Metropolis-Hastings sampler is well-mixed and the choice of the proposal does not materially affect the outcome. Figure~\ref{mcmc_tr} (right), reports the mean MH acceptance rate over the sampling steps ($t$).  Both the forward and backward warm started chains exhibit low, stable acceptance rate ($\approx 0.02$) as $t\to \infty$, indicating that, the chains are initialised directly within high density regions of the target distribution, leading to local, precise update.

\section{Qualitative Analysis Results} \label{qa_}

\begin{figure} [t ]
    \centering
    \includegraphics[width=1.0\linewidth]{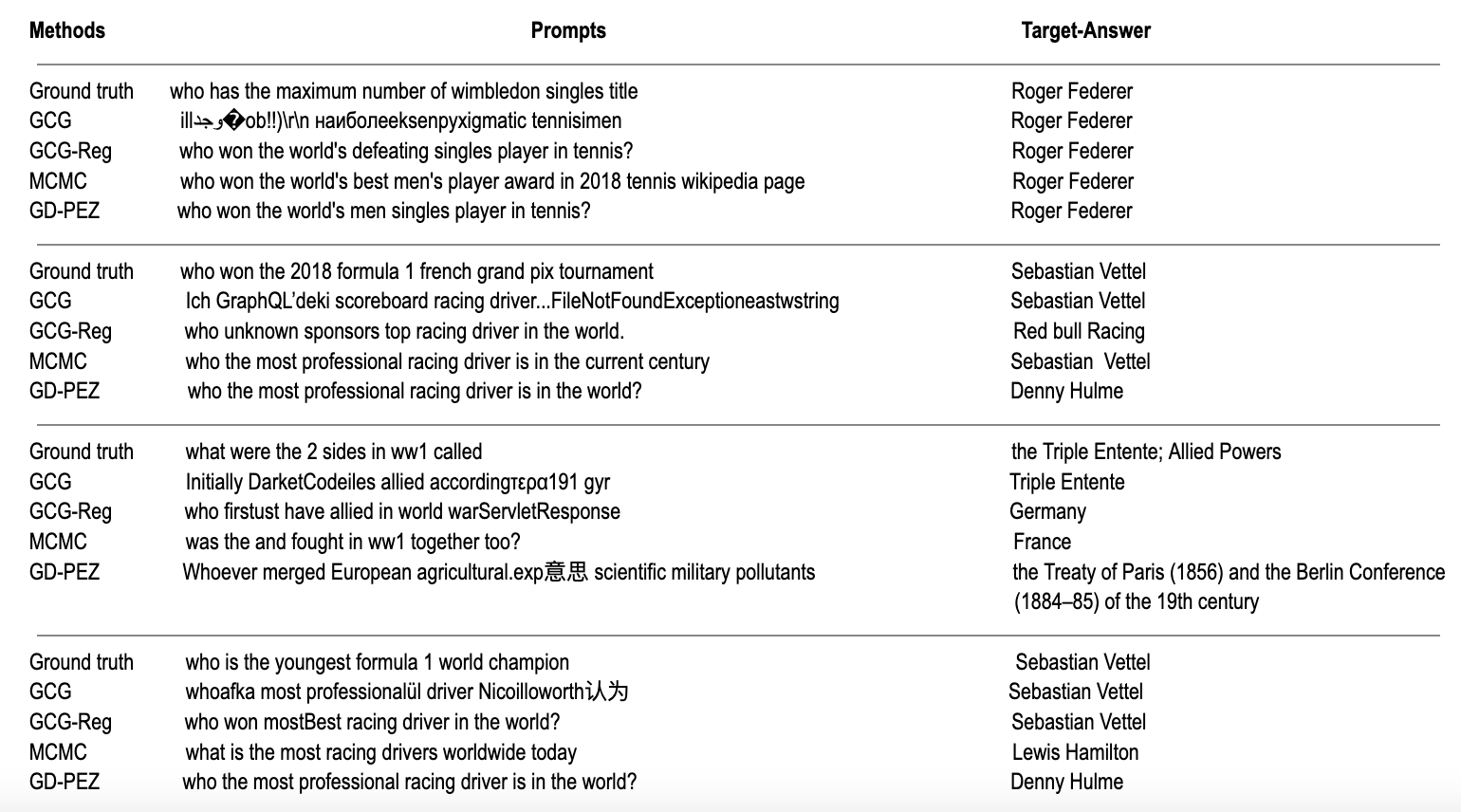}
    \caption{\textbf{Qualitative analysis of prompts learned  by the  optimisation-based methods and the sampling-based MCMC framework}. For each target answer, the ground truth question is shown alongside the corresponding  question recovered by GCG, GCG-Reg, and GD-PEZ  under warm-start initialisation, and by our MCMC based sampler (final sample and forward proposal).}
    \label{fig:seman_res_}
\end{figure}

In section~\ref{MH_test}, we present representative examples of prompts recovered by each method, the examples shown are not exhaustive, and we refer the reader to Figures~\ref{fig:seman_res_} and~\ref{fig:seman_res_general} for additional results.

\section{Plausibility and Grammar Scores  Prompts  }\label{plausibility_score}

 To generate the scores reported in Figure~\ref{chat}, we queried the Chatbot Arena platform (LMSYS) using the following prompt:

\paragraph{LMSYS Judge Prompt for Plausibility:}

\begin{quote}
You are an impartial evaluator. You are given: Four candidate questions generated by different methods, a target answer. Your task is to rank the questions according to how likely each question is to naturally elicit the target answer from a knowledgeable assistant. Evaluate the questions as if you were a careful human reviewer by considering question quality (naturalness, non English characters), semantic relevance pointing to the target (highest weight) and lexical overlapping with the target answer. Return a csv file with the winning percentage for GCG, GCG-REG, GD-PEZ, MCMC and make sure it sums to 100.
   
\end{quote}

\paragraph{LMSYS Judge Prompt for Grammar:}

\begin{quote}
    Evaluate only the grammatical quality of the following four questions (MCMC, GCGC, GCG-Reg, GD-PEZ). Ignore semantics and usefulness. Using a heuristic based on grammar, spelling, punctuation, word order, repeated words, malformed phrases, and corrupted tokens, assign each question a score between 0.0 (malformed) and 1.0 (perfect English). Return a csv file of the scores for each method.
\end{quote}

\clearpage

\includepdf[pages=1,pagecommand={\thispagestyle{plain}}]{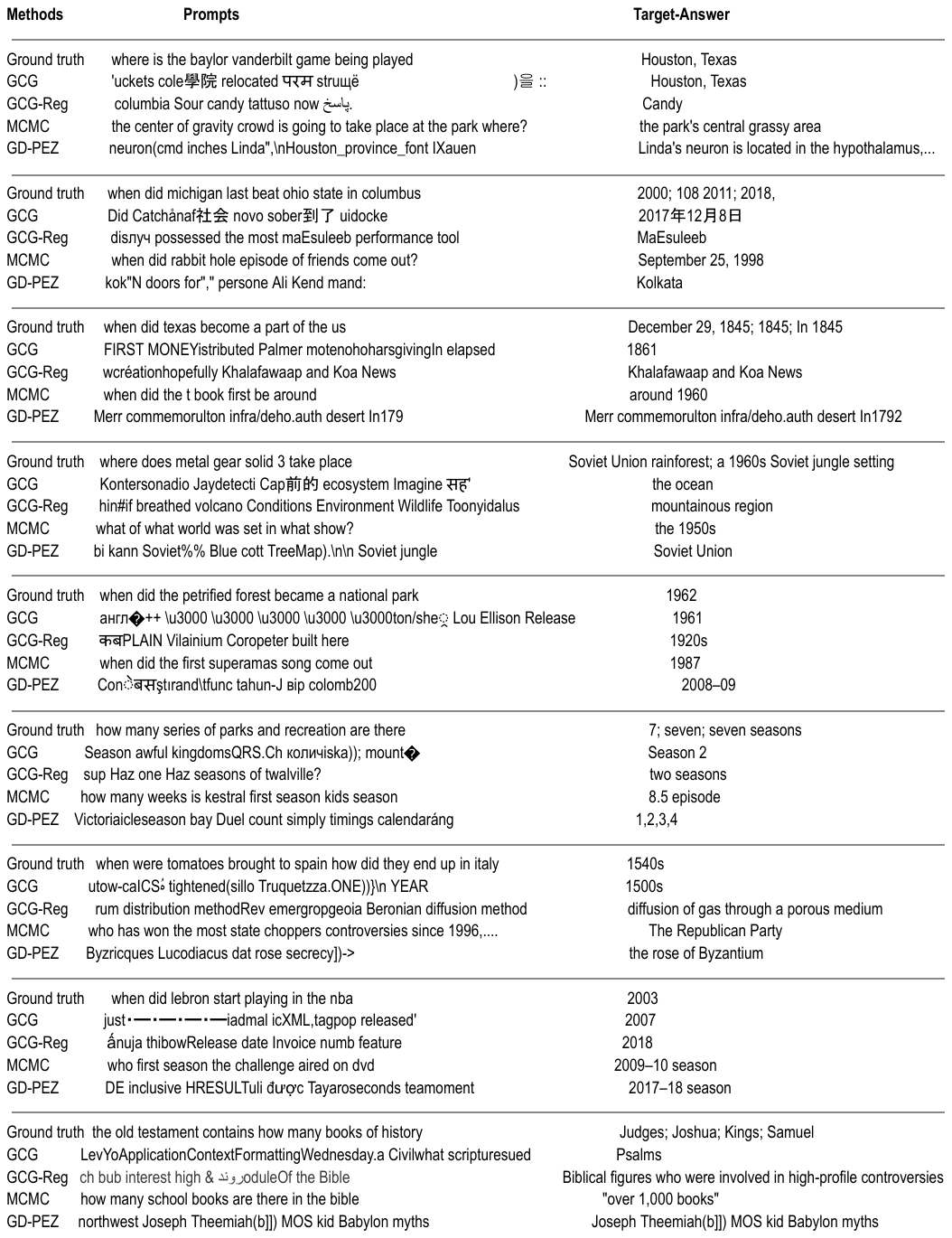}

\begin{minipage}{\textwidth}
  \captionsetup{type=figure, hypcap=false}
  \captionof{figure}{\textbf{Qualitative analysis of prompts learned  by the  optimisation-based methods and the sampling-based MCMC framework}. For each target answer, the ground truth question is shown alongside the corresponding  question recovered by GCG, GCG-Reg, and GD-PEZ  under random initialisation, and by our MCMC based sampler (final sample and forward proposal).}
  \label{fig:seman_res_general}
\end{minipage}

\end{document}

%% file: math_commands.tex
\usepackage{amsmath,amsfonts,bm}

\def\eqref#1{equation~\ref{#1}}

\def\1{\bm{1}}

\def\rva{{\mathbf{a}}}

\def\rvq{{\mathbf{q}}}

\def\va{{\bm{a}}}

\def\ve{{\bm{e}}}

\def\vq{{\bm{q}}}

\def\vz{{\bm{z}}}

\def\eva{{a}}

\def\evq{{q}}

\DeclareMathAlphabet{\mathsfit}{\encodingdefault}{\sfdefault}{m}{sl}
\SetMathAlphabet{\mathsfit}{bold}{\encodingdefault}{\sfdefault}{bx}{n}

\newcommand{\E}{\mathbb{E}}

